\pdfoutput=1
\documentclass[10pt,twocolumn,letterpaper]{article}

\usepackage{3dv}
\usepackage{times}
\usepackage{epsfig}
\usepackage{graphicx}
\usepackage{amsmath}
\usepackage{amssymb}
\usepackage[sort, numbers, square]{natbib}
\usepackage{color}
\usepackage{multirow}
\usepackage{enumitem}
\usepackage{booktabs}
\usepackage{subcaption}
\usepackage[font=footnotesize,labelfont=bf]{caption}
\usepackage[font=scriptsize]{subcaption}
\usepackage[pagebackref=true,breaklinks=true,letterpaper=true,colorlinks,bookmarks=false]{hyperref}
\threedvfinalcopy 

\def\threedvPaperID{219} 

\ifthreedvfinal\fi
\begin{document}

\title{Domain Adaptive 3D Pose Augmentation for In-the-wild Human Mesh Recovery}

\author{Zhenzhen Weng$^1$ \and Kuan-Chieh Wang$^1$ \and Angjoo Kanazawa$^2$ \and Serena Yeung$^1$ \\
\and
${}^{1}$ \text{Stanford University}\\
{\tt\small \{zzweng,wangkua1,syyeung\}@stanford.edu}
\and
${}^{2}$ \text{UC Berkeley} \\
{\tt \small kanazawa@eecs.berkeley.edu}
}

\maketitle
\thispagestyle{empty}

\begin{abstract}

The ability to perceive 3D human bodies from a single image has a multitude of applications ranging from entertainment and robotics to neuroscience and healthcare. 
A fundamental challenge in human mesh recovery is in collecting the ground truth 3D mesh targets required for training, which requires burdensome motion capturing systems and is often limited to indoor laboratories.
As a result, while progress is made on benchmark datasets collected in these restrictive settings, models fail to generalize to real-world ``in-the-wild'' scenarios due to distribution shifts.
We propose Domain Adaptive 3D Pose Augmentation (DAPA) \footnote{Code:  \href{https://github.com/ZZWENG/DAPA_release}{https://github.com/ZZWENG/DAPA\_release}}, a data augmentation method that enhances the model's generalization ability in in-the-wild scenarios.
DAPA combines the strength of methods based on synthetic datasets by getting direct supervision from the synthesized meshes, and domain adaptation methods by using ground truth 2D keypoints from the target dataset. 
We show quantitatively that finetuning with DAPA effectively improves results on benchmarks 3DPW~\citep{von2018recovering} and AGORA~\citep{patel2021agora}. We further demonstrate the utility of DAPA on a challenging dataset curated from videos of real-world parent-child interaction.
\end{abstract}

\section{Introduction}
\label{sec:intro}

\begin{figure}[t]
\centering
\includegraphics[width=\columnwidth,page=1]{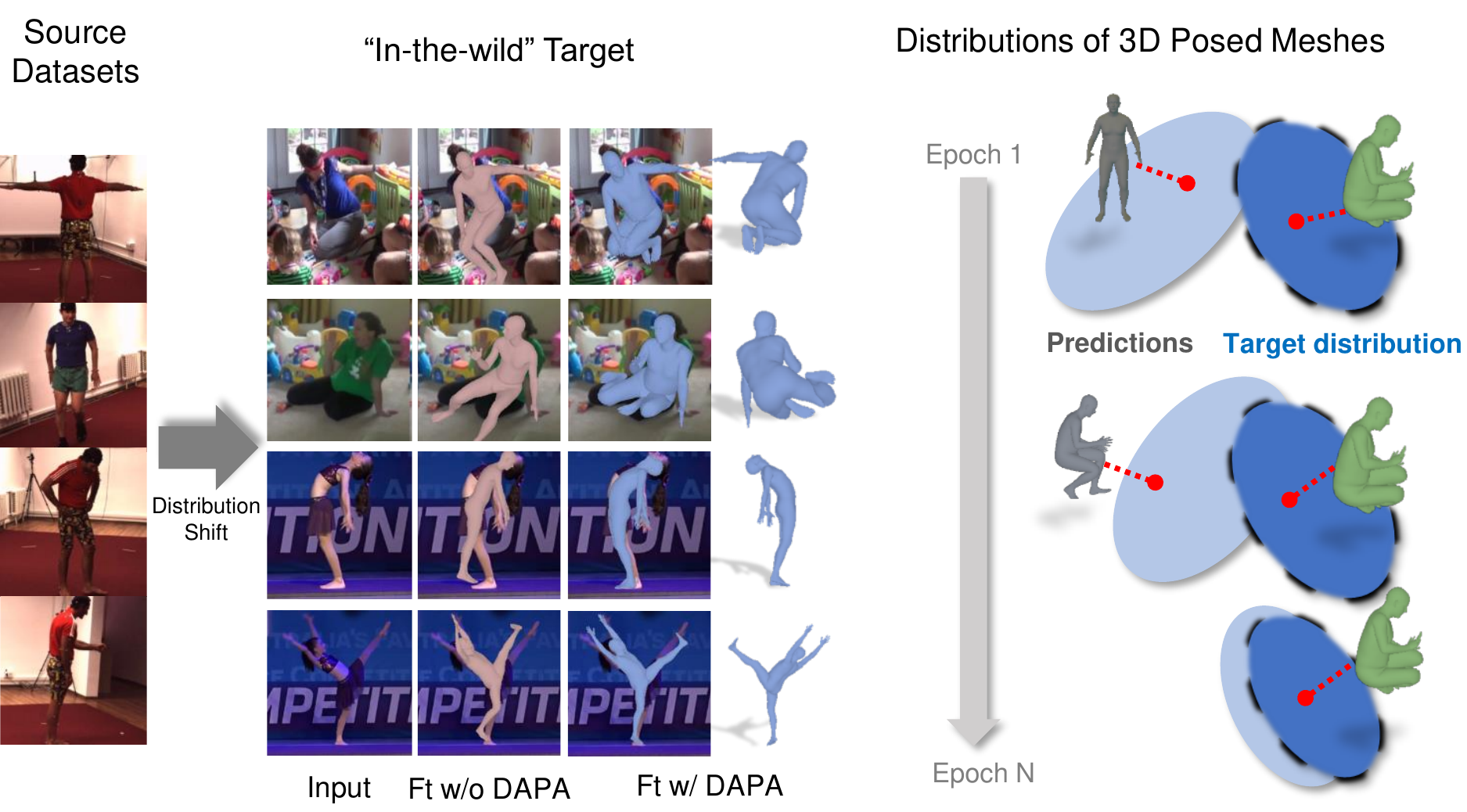}
\caption{\textit{Left - } Distribution shift makes human mesh recovery on in-the-wild datasets particularly challenging. Current fine-tuning methods (`Ft w/o DAPA') are often biased towards poses in the source distribution. \textit{Right - } Our method DAPA progressively adapts to the target distribution of poses. }
\label{fig:fig1}
\end{figure}

Human mesh reconstruction from single-view images \cite{bogo2016keep,kanazawa2018end,weng2021holistic} brings a lot of exciting opportunities in domains such as VR/AR \cite{zhang2021vid2player,zhu2020reconstructing}, healthcare \cite{clever2020bodies} and autonomous driving \cite{yang2021s3}.
Yet, a fundamental challenge in 3D human recognition is in collecting the groundtruths for training, which often requires a motion capturing (MoCap) system on the target and is not feasible in scenarios such as parents having natural interactions with infants or athletes making extreme movements. To realize these transformative applications, we need methods that can adapt models trained on standard datasets collected in studios to a given target dataset.

To illustrate the callenge of distribution shifts, constrast images of people captured in the standard MoCap setting to images of parents attending to infants (Fig.~\ref{fig:fig1}-\emph{left}). 
In these datasets, not only are the image appearances different, but the distributions of the poses are also different. 
Parents are often sitting or lying on the ground when interacting with children, but individuals in MoCap datasets are often standing. This is only one example of the more ubiquitous problem of generalizing to arbitrary in-the-wild target datasets. 
A solution needs to be aware of the target distribution, and adapt a model specifically given a different target dataset.

In this work, we propose a method that can adapt an existing single-view human mesh reconstruction model to challenging datasets with vastly different appearance and pose distribution using only 2D keypoint supervision, which is often the only supervision obtainable in the wild. 
The idea is that from a pretrained model, we can get a rough estimate of the poses in the target dataset despite some inaccuracies caused by domain shift. Then, by augmenting these estimated poses using a realistic pose prior and rendering pipeline, the augmented pose and image pairs can provide strong 3D supervision not available in the original target dataset. Using this supervision, the model's performance improves around the neighborhood of the augmented poses. The next time the model encounters a real image from the target distribution in this neighborhood, the estimated pose improves.
Gradually, with the guidance of 2D supervision on the real samples, both the predicted pose distributions and augmented distributions shift over time to match that of target pose distribution (Fig.~\ref{fig:fig1}-\emph{right}). 

In summary, our contributions are as follows.
\begin{itemize}[noitemsep,topsep=0pt]
\item[$\bullet$] We propose a method for closing the domain gap in single view human mesh reconstruction, which we name Domain Adaptive Pose Augmentation (DAPA). 
DAPA is \emph{backbone-agnostic} - it takes the pose prediction of the backbone and generates synthetic data on-the-fly in an adaptive way. In this paper, we showcase the effectiveness of DAPA using the backbone of \citet{kolotouros2019learning}.
\item[$\bullet$] On 3D benchmarks 3DPW~\citep{von2018recovering} and AGORA \cite{patel2021agora}, we show that finetuning with 2D keypoints and additional synthetic examples generated by DAPA gives better performance than existing finetuning methods \cite{kolotouros2019learning,joo2020exemplar}. In addition, our method achieves competitive results compared to finetuning using 3D annotations.
\item[$\bullet$] We showcase our method's capability in closing significant domain gap on a challenging dataset SEEDLingS \cite{bergelson2019day}.
Fine-tuning with only noisy 2D keypoints obtained by OpenPose~\cite{cao2019openpose}, DAPA improves over finetuned baselines by a large margin on the challenging poses. We additionally demonstrate DAPA's capability in recovering extreme acrobatic poses through qualitative results on a YouTube gymnastics video.
\end{itemize}


\section{Related Work}
\label{sec:relatedworks}
There are a variety of approaches to improve generalization of human mesh recovery models for in-the-wild data. A useful dichotomy we can consider is between approaches that aim to improve the overall generalization, and approaches that are aware of the target distribution (i.e., domain adaptive approaches). The first type of approaches include methods that perform joint training with auxiliary in-the-wild or synthetic datasets, and methods that utilize data augmentation. The second type of approaches (i.e. domain adaptation approaches) can be further broken down by supervision type on the target datasets -- ground truth 2D/3D keypoints, or detected 2D keypoints. Our work falls in the category of domain adaptation methods using ground truth or detected 2D keypoints as supervision.

\subsection{Improving Overall Human Mesh Recovery Generalization.}
\paragraph{Train with in-the-wild 2D Datasets.}
Due to the lack of in-the-wild 3D datasets, one approach to address in-the-wild distribution has been to jointly train on indoor MoCap datasets as well as a variety of challenging 2D datasets (e.g. COCO, MPI-II)~\cite{kanazawa2018end,kolotouros2019learning,habibie2019wild}. 
An obvious drawback of these approaches is that they rely on the quality of the auxiliary datasets, and how much these datasets overlap with the given target dataset.  

\vspace{-\baselineskip}
\paragraph{Train with Synthetic Datasets.}
Previous works have explored using synthetic images to improve generalization of human pose estimation. \citet{varol2017learning} and \citet{chen2016synthesizing} build large synthetic training sets by rendering posed humans with graphic engines. 
The poses are drawn from the 3D ground truths of the existing lab-captured datasets.
More recently, new methods for synthesizing datasets for realistic images with posed humans, take into account physical contact and interactions~\citep{Hassan:CVPR:2021,PSI:2019}. 
\citet{Patel:CVPR:2021} and \citet{kocabas2021spec} curate large synthetic datasets by putting 3D commercial human body scans into 3D scenes and taking snapshots of the populated scenes. 
The intuition is that, to recover in-the-wild human meshes, a large and diverse training set can be built to cover all possible target poses.
This idea seems inefficient if our goal is to do well on a given target domain, and does not provide guarantees on whether a target distribution is covered in the precomputed training set. 
This type of approach can also be thought of as orthogonal to domain adaptation approaches (including ours) since the generalist model can still benefit from being adapted to a given target dataset.
\vspace{-\baselineskip}
\paragraph{Data Augmentation for 3D Skeleton Estimation.}
A closely relevant research topic to HMR is 3D human keypoints (i.e. skeleton) detection.  
To address the lack of 3D supervision in in-the-wild datasets, previous 3D human skeleton works have used data augmentation to improve model generalization. They augment data by stitching image patches \cite{sarandi2018robust}, augmenting the keypoint heatmaps \cite{cheng20203d}, or adding perturbation to the ground truth 3D skeletons \cite{gong2021poseaug}. Recent 3D human mesh recovery works train on cropped \cite{biggs20203d,rockwell2020full,joo2020exemplar} images to boost the robustness of the models in case of occlusion. 
Our work takes inspiration from the success of data augmentation methods \cite{gong2021poseaug} in the skeleton detection problem and extend them to the HMR setting. 
Compared to 3D skeleton detection, HMR requires a more complete parametrization of the human body
which are lacking in the 3D skeleton methods.
Our work applies more significant augmentation to the training set, by not simply altering the appearances of the RGB images, but also re-posing the humans and rendering a new image given the reposed human, not achievable if only skeleton is considered.

\subsection{Domain Adaptive Methods.}
In closing the domain gap, works like EFT \cite{joo2020exemplar} and SPIN \cite{kolotouros2019learning} can be used to fit to the target 2D datasets using pretrained priors. EFT \cite{joo2020exemplar} proposed to use a pretrained model as pose prior to curate pseudo ground truth meshes for those 2D datasets, which can then be used for finetuning, or joint training with 3D datasets to improve HMR model generalization. SPIN \citep{kolotouros2019learning} leverages the benefit of optimization-based approaches \cite{lassner2017unite,bogo2016keep,SMPL-X:2019} - in each training iteration, an optimization routine takes the initial estimates of the body model (i.e. SMPL \cite{loper2015smpl}) parameters and then iteratively optimizes the parameters to minimize the 2D reprojection loss as well as the discrepancy to a predefined pose prior (a mixture model). The output of the optimization routine then provides additional model-based supervision for that iteration.
Unlike EFT and SPIN that utilize fixed pose priors, we instead supplement the training set with automatically generated synthetic data with poses that are representative of the target domain.

Another type of approach directly adapt an existing model on a target dataset. Recently, \citet{guan2021bilevel} proposed an unsupervised online adaptation method for human mesh recovery on streaming videos leveraging 2D detections and temporal consistency. Our method does not use temporal information and works with single view images. 

\section{Method}
\label{sec:method}
\begin{figure*}[t]
\centering
\includegraphics[width=\textwidth,page=17, trim=0 380 150 0, clip]{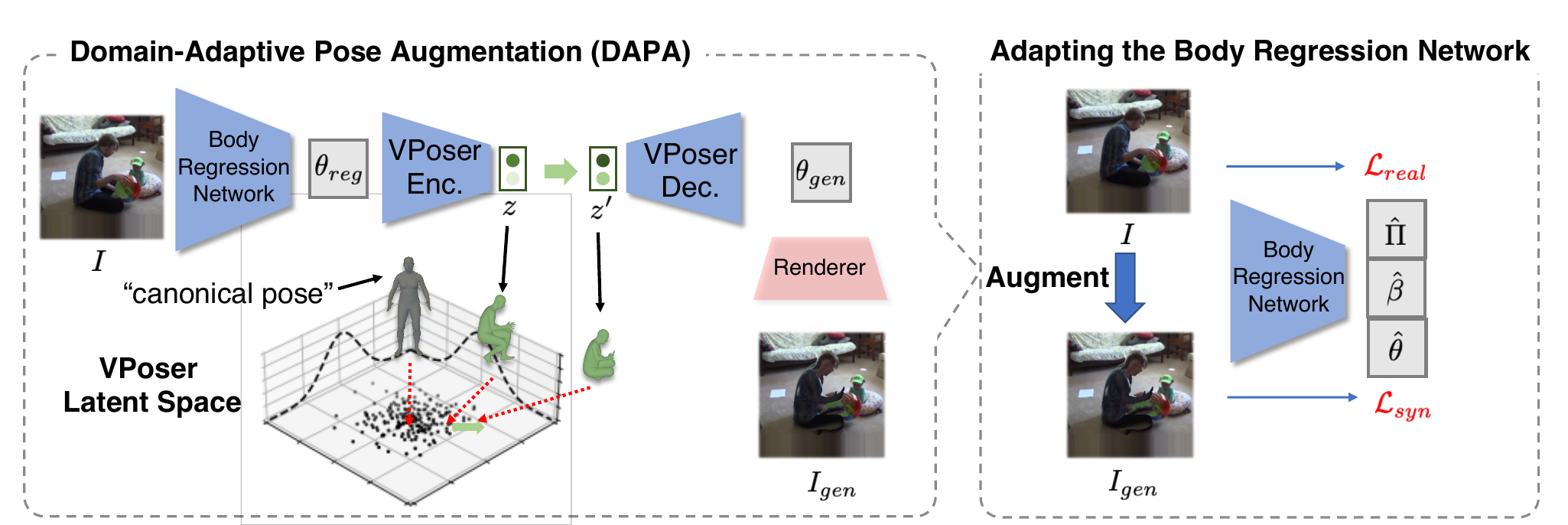}
  \caption{We propose domain-adaptive pose augmentation (DAPA) to combat the distribution shift problem in in-the-wild mesh reconstruction.  DAPA augments the estimated pose given an target image by moving it further away from the canonical pose in the latent space of VPoser, a pretrained pose prior. Then, the body regression network is finetuned on both the real target images, and their augmented counterparts.}
  \label{fig2}
\end{figure*}
\paragraph{Problem Setup.}
We focus on the problem of adapting pretrained 3D human mesh reconstruction models to new datasets, especially in-the-wild target datasets where no 3D ground truths are available for training, and the pose distribution is different from the source (i.e., indoor lab environments).  The model is pretrained on source dataset $D_{src} = \{I^n, X^n \}_{n=1}^{N_{src}}$ with pairs of images $I \in \mathbb{R}^{h\times w \times 3}$ and 3D ground truth annotations of $k$ body joints $X \in \mathbb{R}^{k\times 3}$ for each person. Our goal is to secure a good accuracy on the target dataset $D_{tar} = \{I^n, J^n\}_{n=1}^{N_{tar}}$, using only 2D keypoint annotations $J \in \mathbb{R}^{k\times 2}$ in the target domain.
\vspace{-\baselineskip}
\paragraph{Overview.}
An overview of our method is outlined in Figure \ref{fig2}.
The key contribution of our method is the Domain Adaptive Pose Augmentation (DAPA) module that can generate pairs of images and ground truth poses for supplementing adaptation of a pretrained body regression network on the target dataset.  
A body regression network, such as HMR~\cite{kanazawa2018end} or SPIN~\cite{kolotouros2019learning}, is first pretrained on the source.
During the adaptation stage, our full method works as follows: 1. Given an input image in the target domain, a body regression network provides an initial estimated pose, 2. DAPA takes in the initial pose, and produces an augmented pose and renders an image of a person with that augmented pose, 3. Finally, the body regression network is finetuned on both the real input image and the augmented pair. At test-time, we use a single forward pass of the adapted body regression network. In the following sections, we describe the body regression network (Section~\ref{sec:method:mesh}), our data augmentation module, DAPA, (Section~\ref{sec:method:dapis}), and the losses used for finetuning (Section~\ref{sec:method:mesh-estimator}).

\subsection{Body Regression Network.}
\label{sec:method:mesh}
\paragraph{Human Body Representation.}
We use the Skinned Multi-Person Linear (SMPL) model \cite{loper2015smpl} to represent the 3D mesh of the human body. The SMPL body model is a differentiable function $\mathcal{M}(\theta, \beta)$ that takes a pose parameter $\theta \in \mathbb{R}^{69}$ and shape parameter $\beta \in \mathbb{R}^{10}$, and returns the body mesh $M$ with $6890$ vertices. A linear regressor $W$ can be pretrained to get the major body joints $X \in \mathbb{R}^{k\times 3} = WM$, which are a linear combination of the mesh vertices. 
\vspace{-\baselineskip}
\paragraph{Body Regression Network.} 
The body regression network is a function, denoted by $f$, that takes an input RGB crop $I$ centered on a human, and predicts the body shape $\beta_{reg}$, pose $\theta_{reg}$, and camera parameters $\Pi_{reg}$: 
$$\{\beta_{reg}, \theta_{reg}, \Pi_{reg}\} = f(I)$$

From these parameters we can get the joints in 3D, $X_{reg} = W \mathcal{M}(\theta_{reg}, \beta_{reg})$, as well as 2D projections of the joints, $J_{reg} = \Pi_{reg}(X_{reg})$. 
We use the the same network architecture as SPIN~\cite{kolotouros2019learning}. 
SPIN uses a body regression network to initialize the estimated mesh, and then performs iterative refinement to optimize the mesh by minimizing 2D keypoint re-projection loss. We follow SPIN's pretraining process, and pretrain the body regression network on the source datasets using the losses proposed in SPIN. 
Our method differs from SPIN during the adaptation stage.

\subsection{Domain Adaptive Pose Augmentation}

\label{sec:method:dapis}
\paragraph{Pose Augmentation.}
Given in-the-wild target images, poses estimated using the pretrained body regression network often suffer from being biased towards poses in the source datasets. 
This is illustrated in Figure~\ref{fig:fig1} where the input image contains a human sitting on their legs, but the estimated pose has only half-bent knees due to the bias of seeing mostly poses with straight legs. 
Motivated by this recurring issue, we aim to generate synthetic pose and image pairs that are closer to the target pose distributions.  
Our domain adaptive pose augmentation (DAPA) module takes as input an estimated pose from the body regression network, and perturbs the estimated pose in a realistic pose space. This allows it to move away from the canonical pose (i.e., a standing pose with mostly straight limbs). 

Our approach uses VPoser~\cite{SMPL-X:2019}, which is pretrained on AMASS~\cite{mahmood2019amass}, a massive dataset of real human poses including rare and extreme poses. 
VPoser is trained as a Variational Autoencoder (VAE) whose observation variable is human pose in terms of SMPL pose parameter. 
After being trained on AMASS, VPoser effectively learns the distribution of realistic human poses and can be used as a human pose prior.
Our augmentation strategy embeds the input pose into the latent space of VPoser, and applies a multiplicative noise greater than $1$ before decoding the augmented latent pose.  
The intuition is that, in the latent space of VPoser, less frequent and more challenging poses are embedded further away from the origin due to their lesser likelihood of appearing in a dataset like AMASS (See Figure ~\ref{fig2}).
This formulation ensures the augmented pose is realistic because of our usage of a pretrained pose prior, and relevant to the target dataset because the input is from the estimated pose on an image from the target dataset. 
In the early phases of this work, we considered other formulations such as learning a Generative Adversarial Network (GAN) to capture the distribution of the target pose, but this was challenging as no ground truth poses are available in the target dataset.

Our augmentation strategy is described as follows:
\begin{align}
\mu, \sigma &=\mathrm{Encoder}_{\mathrm{VPoser}}(\theta_{reg}) \\
z &\sim \mathcal{N}(\mu, \sigma I) \\
\tilde{z} &= z \odot (1 + s \epsilon), \quad \epsilon \sim \mathcal{U}[0,1] \label{eq:vposer} \\
\theta_{syn} &= \mathrm{Decoder}_{\mathrm{VPoser}} (\tilde{z})
\end{align}

where $s$ is a constant scalar, and $\epsilon$ is from a multivariate uniform distribution of the same dimension the VPoser latent space. $\odot$ denotes the Hadamard product. 

There are two driving forces to encourage synthesizing poses in the target distribution: 1) using the predicted pose given an image in the target distribution as input to VPoser, and 2) the multiplicative noise in the latent pose space.  Since this augmentation is done on-the-fly, our method benefits from the symbiosis that as the body regression network improves during finetuning, the input to the VPoser distribution also becomes closer to the target distribution. 
\vspace{-\baselineskip}
\paragraph{Synthetic Image Rendering.}
The sampled poses $\theta_{syn}$ are then rendered as RGB images so that we have those synthetic training images with ground truths as an additional source of training supervision in our adaptation stage. To synthesize an image given an augmented pose as described in the previous section, we use a texture model, which takes in a pose and a texture, and creates a textured human. Specifically, as in \citet{kanazawa2018learning}, our texture model takes the image feature (the output of the ResNet in the body regression network) and predicts a flow map that specifies where each pixel in the input image $I$ should go into the UV texture $I^{uv}$. We can then texture the posed SMPL bodies and render using Neural Mesh Renderer \cite{kato2018neural}.
The texture model is pretrained on the source, and frozen during adaptation.

\subsection{Adaptation Framework}
\label{sec:method:mesh-estimator}
During finetuning on the target dataset, each batch of real input images is supplemented by the same number of synthetic images using DAPA (Section~\ref{sec:method:dapis}). At each training iteration, the body regression network $f$ is updated to minimize the overall loss function:
\begin{align}
    \mathcal{L} &= \mathcal{L}_{real} + \mathcal{L}_{syn}  \label{eq:real_plus_syn}
\end{align}
where $\mathcal{L}_{real}$ and $ \mathcal{L}_{syn}$ denotes losses for the real and synthesized data respectively.
For real data, we minimize the 2D re-projection loss 
$\mathcal{L}_{real} = \lambda_{2D} ||J_{reg} - J_{gt}||, \label{eq:real_loss}$
where $J_{reg}$ denotes the predicted/regressed 2D keypoints, and $J_{gt}$ the ground truth 2D keypoints.
Note, we focus on the setting where no 3D ground truth keypoints are available, as this is often the case for in-the-wild datasets.

For synthetic training examples, we have the advantage of having access to the ground truth SMPL parameters $\{\theta_{syn}, \beta_{syn}, \Pi_{syn} \}$ as well as the corresponding 3D joints $X_{syn}$ and 2D projections $J_{syn}$. Hence, for the synthesized data, the loss function can be much more informative and takes on the following form:
\begin{align}
    \mathcal{L}_{syn} &= \lambda_{2D} \mathcal{L}_{syn,2D} + \lambda_{3D} \mathcal{L}_{syn,3D} \nonumber \\
    & + \lambda_{\theta} \mathcal{L}_{syn,\theta} + \lambda_{\beta} \mathcal{L}_{syn,\beta}  \label{eq:syn_loss}
\end{align}
where
\begin{align}
\mathcal{L}_{syn,2D} &= ||J_{syn,reg} - J_{syn}|| \nonumber \\
\mathcal{L}_{syn,3D} &= ||X_{syn,reg} - X_{syn}|| \nonumber \\
\mathcal{L}_{syn,\theta} &= ||\theta_{syn,reg} - \theta_{syn}|| \nonumber \\
\mathcal{L}_{syn,\beta} &= ||\beta_{syn,reg} - \beta_{syn}|| \nonumber
\end{align}
and $\lambda$s are hyper-parameters. In other words, with ground truth SMPL parameters, we can directly supervise the predicted 2D/3D keypoints as well as the SMPL parameters.

\section{Experiments}
\label{sec:experiments}
\subsection{Overview}
\label{sec:experiments_overview}
We present results on applying our method, DAPA, to the task of weakly-supervised domain adaptation for human mesh recovery. 
We showcase the effectiveness of DAPA on three datasets: 3DPW~\cite{von2018recovering} and AGORA~\cite{patel2021agora}, 3D benchmarks where ground truth mesh is available for evaluation, and SEEDLingS~\cite{bergelson2019day}, a dataset derived from real-world videos of parent-child interaction. \emph{We note that datasets like SEEDLingS is our target. For 3DPW and AGORA, the domain difference is less severe, still we conduct the experiment for completeness, where our approach still shows meaningful improvements.}

We also note that DAPA is backbone-agnostic. While it is possible to apply DAPA to more recent models \cite{lin2021mesh,Kocabas_PARE_2021} that achieve state-of-the-art performance on 3DPW and AGORA, those models would require ground truth 3D supervision from the target dataset if used for finetuning. We consider the setting where only 2D keypoints from the target dataset are available during adaptation, which is often the case in real-world applications. To that end, we choose to adopt the backbone of SPIN~\citep{kolotouros2019learning}, which allows us to make fair comparison to existing finetuning methods \cite{kolotouros2019learning,joo2020exemplar} under the same setting.

\subsection{Target Datasets}
\paragraph{3DPW \cite{von2018recovering}} consists of video sequences captured in mostly outdoor conditions where human subjects are performing tasks such as walking or chasing the bus. IMU sensors were used to obtain ground truth 3D SMPL parameters and mesh annotations. 
The dataset contains $22,735$ persons in the training set and $35,515$ persons in the test set. We only use the OpenPose detections in our experiments.
\vspace{-\baselineskip}
\paragraph{AGORA~\cite{patel2021agora}} is a recent synthetic dataset that is constructed using commercially-available and high-resolution 3D body scans. The dataset includes challenging cases such as environmental occlusion and person-person occlusion, which makes it useful for evaluating generalization ability to in-the-wild images. AGORA contains $14,529$ training and $1,225$ validation images with multiple persons in each image. 
Note that we use ground truth 2D keypoints from AGORA to adapt the pretrained models.
\vspace{-\baselineskip}
\paragraph{SEEDLingS (Study of Environmental Effects on Developing LINGuistic Skills) \cite{bergelson2019day}} is a large parent-child interaction dataset with monthly video recordings of $46$ participants. The dataset is released via Databrary \cite{Databrary} and the majority of the recordings are publicly accessible. Due to the nature of this dataset, the adult poses are mostly kneeling and sitting in various ways, which makes it an ideal dataset to validate our method's effectiveness in adapting to a target domain. We sample frames from the video recordings to construct training and test sets. The training set consists of $5,631$ persons and the test set of $450$ persons. We do not have access to ground truth 2D keypoints on the training set. Instead, we use OpenPose \cite{cao2019openpose} keypoint detections as noisy ground truth during training. This is a representative setting of deploying pretrained human mesh recovery models on real-world data without any manual annotations. 
Pre-processing details are in Supplementary. 
To evaluate our method, we collected ground truth 2D keypoints of the $450$ adults in the test set. We will release our annotations.

\subsection{Implementation Details}
As discussed in Sec. \ref{sec:experiments_overview}, we adopt the backbone of SPIN~\citep{kolotouros2019learning} as the pose regression network. The encoder is a ResNet-50 \cite{he2016deep} followed by fully-connected layers that iteratively regress SMPL parameters.
We pretrain our pose regression network and texture model on the datasets used in SPIN (i.e. Human3.6M \cite{ionescu2013human3}, MPI-INF-3DHP \cite{mehta2017monocular}, LSP \cite{johnson2010clustered}, LSP-Extended \cite{johnson2011learning}, COCO \cite{lin2014microsoft}, MPII \cite{andriluka20142d}). 
The hyper-parameters in Eq. \ref{eq:real_loss} and \ref{eq:syn_loss} are the default parameters used in SPIN, namely $\lambda_{2D} = \lambda_{3D} = 5$, $\lambda_{\theta}=1$, $\lambda_{\beta}=0.001$. 
VPoser \cite{SMPL-X:2019} is pretrained on AMASS \cite{mahmood2019amass}.  
$s$ in Eq. \ref{eq:vposer} is $0.1$ for 3DPW and $0.5$ for other datasets. Models are trained until the loss curves plateau. 

\subsection{Results}
In this section, we perform finetuning experiments on all three datasets where we fine tune a pretrained model on an unlabelled training set from target domain, and evaluate on a test set from target domain. We then show on 3DPW that DAPA can be used as a test optimization method where we finetune a pretrained model and evaluate on a test set from target domain. Lastly, we demonstrate the utility of DAPA on a challenging real-world dataset SEEDLingS. To further demonstrate DAPA's strength, we additionally show results on a difficult gymnastics video with extreme poses.

\begin{table}[h!]

\centering
\scriptsize

\resizebox{\columnwidth}{!}{\begin{tabular}[t]{@{} l *5c @{}}
\toprule
 \multicolumn{1}{c}{Method} & {Supervision}~ & MPJPE & Rec Err & Vertex Err &  PA Vertex Err \\ 
\midrule
SPIN-pt & {pt} & 96.9 & 59.6  & 172.7 & \textbf{110.7} \\
\midrule
SPIN-ft &  {ft w/ 2D} &  99.4 & 64.1 & 161.9 & 114.3  \\
EFT-ft & {ft w/ 2D} & 102.5 & 60.3 & 166.4 & 113.1 \\ 
DAPA (Ours) \\
\;\;- $\mathcal{L}_{real}$ only & {ft w/ 2D} & 138.7 & 59.4 & 247.7 & 200.4\\
\;\;- $\mathcal{L}_{syn}$ only & {ft w/ 2D} & 99.4 & 62.3 & 172.8 & 123.0 \\
\;\;- 0 perturb & {ft w/ 2D} & 96.2 &  59.9 & 167.2 & 119.8 \\
\;\;- Rand. pose & {ft w/ 2D} & 97.8 & 61.0 & 168.6 & 121.9\\
\;\;- Full model & {ft w/ 2D} & \textbf{94.5} & \textbf{59.4} & \textbf{160.2} & 112.7 \\
\bottomrule
\end{tabular}}
\caption{Finetuning results on 3DPW.}
\label{table:main_table_a}
\resizebox{\columnwidth}{!}{\begin{tabular}[t]{@{} l *6c @{}}
\toprule
\multicolumn{1}{c}{Method} & {Supervision} & MPJPE & MVE  & NMJE & NMVE  & F1 \\ 
\midrule
 SPIN-pt & {pt} & 175.1 & 168.7 & 223.1 & 216.3 & 0.78 \\ 
 \midrule
 SPIN-ft \\
\;\; -EFT & {ft w/ 3D} & 171.5 & 165.8 & 219.9 & 212.6 & 0.78 \\
\;\; -AGORA & {ft w/ 3D} & 153.4 & 148.9 & 199.2 & 193.4 & 0.77 \\
\midrule
SPIN-ft \\
\;\; -AGORA-2D & {ft w/ 2D} & 202.2 & 201.5 & 266.1 & 265.1 & 0.76 \\ 
DAPA (Ours) & {ft w/ 2D} & \textbf{168.3} & \textbf{164.3} & \textbf{218.6} & \textbf{213.4} & \textbf{0.77}  \\ \bottomrule
\end{tabular}}
\caption{Finetuning results on AGORA.}
\vspace{4px}
\label{table:main_table_aa}
\resizebox{\columnwidth}{!}{\begin{tabular}{@{} l *5c @{}}
\toprule
 Method & MPJPE & Rec Err & Vertex Err & PA Vertex Err\\ 
\midrule
SPIN &  94.73 & 61.23 & 153.60 & 107.3 \\
EFT & 93.38 & 56.58 & 150.21 & \textbf{104.28}
 \\ 
BOA \citep{guan2021bilevel} (eval at the end) & 136.80 & 79.18 & - & - \\ 
DAPA (Ours) & \textbf{81.96} & \textbf{53.39} & \textbf{147.19} & 111.23 \\
\bottomrule
\end{tabular}}
\caption{Test optimization results on 3DPW}
\label{table:main_table_c}

\caption{Quantitative results on 3DPW and AGORA.
}
\label{table:main_table}
\end{table}

In Table \ref{table:main_table} we report finetuning results on 3DPW and AGORA as well as test optimization results on 3DPW. Next, we elaborate on the experiments protocols.
\vspace{-\baselineskip}
\subsubsection{Finetuning experiments on 3DPW and AGORA}
\paragraph{3DPW.} In Table \ref{table:main_table_a} and \ref{table:main_table_aa}, we compare DAPA to previous finetuning methods. Following previous works we report Mean Per Joint Position Error (MPJPE), Reconstruction Error, Vertex Error and Procrustes-aligned Vertex Error. SPIN-pt is the pretrained model that was trained on datasets (Human3.6M, MPI-INF-3DHP, LSP, LSP-ext, COCO, MPII) without 3DPW-train. We then finetune SPIN-pt on 3DPW-train and compare to existing finetuning methods SPIN \citep{kolotouros2019learning} and EFT \citep{joo2020exemplar}
In SPIN-ft, we finetune the pretrained SPIN-pt with the SPIN framework and only use OpenPose keypoints as 2D supervision. In EFT-ft, we use EFT to obtain pseudo ground truth mesh for every image in 3DPW-train, and then finetune SPIN-pt using the augmented training set. We then use DAPA to finetune the model and additionally include four ablated versions of DAPA where we use either loss term in Eq. \ref{eq:real_plus_syn}, add 0 perturbation in pose sampling (``0 perturb"), and sample random poses with VPoser (``Rand. pose"). We show that the proposed perturbed adaptive sampling is indeed necessary in achieving the desired performance. We also observe that DAPA achieves better results than SPIN and EFT on all metrics, and hightlight its significant improvement on MPJPE and Vertex Err. 
\vspace{-\baselineskip}
\paragraph{AGORA.} In Table \ref{table:main_table_c}, we report finetuning results on AGORA. As in \citet{patel2021agora}, we report Normalized Mean Joint Error (NMJE), Mean Vertex Error (MVE), Normalized Mean Vertex Error (NMVE), and Normalized Mean Joint Error (NMJE), as well as the F1 score. The metrics account for human detection performance, i.e. the commonly used pose estimation metrics such as MPJPE are normalized by the standard detection metric, F1-score (the harmonic mean of recall and precision). During evaluation, the estimated persons are first matched to the ground truth persons in AGORA based on 2D joint error. The body metrics are then computed on the matched predictions. The baselines are different in terms of their training schemes and supervision types. Model SPIN-pt is pretrained on the datasets in SPIN (i.e. source datasets), and is equivalent to the model after our pretraining stage. SPIN-ft-EFT and SPIN-ft-AGORA obtain improvement by finetuning with 3D supervision on auxiliary datasets, EFT([MPII+LSPet+COCO]) and AGORA respectively. Relative to these, in our setting when only 2D keypoint annotations are available, finetuning on AGORA using the SPIN method (SPIN-ft-AGORA-2D) performs even worse than the pretrained SPIN-pt on the test set. In contrast, our proposed DAPA, which uses adaptively-generated synthetic images instead of SPIN's intermediate optimization routine, significantly outperforms SPIN-ft-AGORA-2D, and also outperforms SPIN-pt by a good margin. In addition, we outperform SPIN-ft-EFT that is finetuned on auxiliary 3D datasets.

\begin{figure*}[t!]
\centering
\includegraphics[width=0.9\textwidth,page=3]{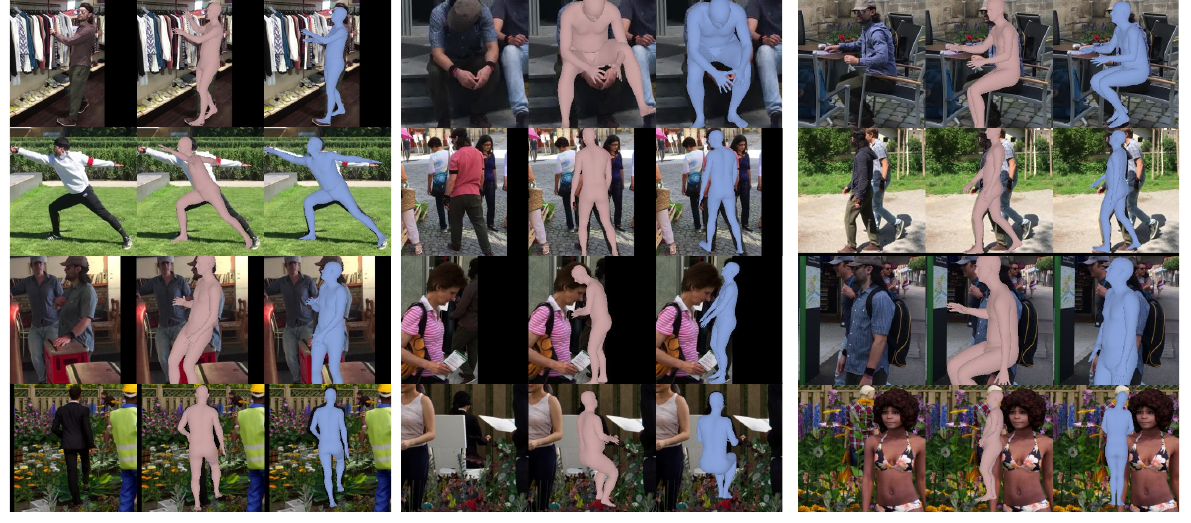}
\caption{Qualitative results on 3DPW (top 3 rows) and AGORA (bottom row). Each column, left to right: input, SPIN-ft result, DAPA (Ours) result.}
\label{fig:3dpw_agora_qualitative}
\end{figure*}

In Figure \ref{fig:3dpw_agora_qualitative}, we include qualitative comparison of DAPA and baseline methods, and show that the reconstruction quality using DAPA is overall better. Additional qualitative examples are included in the Supplementary.
\vspace{-\baselineskip}

\subsubsection{Test optimization experiments on 3DPW}
In Table \ref{table:main_table_c} we show that DAPA can be used in test optimization. In this setting, we finetune the pretrained model on 3DPW-test (without ground truths) and evaluate on the same test set. For comparison, we finetune the pretrained model with SPIN using only OpenPose keypoints as supervision. For the EFT comparison, we use EFT to curate pseudo ground truth mesh for each image in 3DPW-test, and then finetune using the augmented test set. In addition, we compare to the frame-based version of BOA \cite{guan2021bilevel}, an online domain adaptation framework on streaming frames. For fair comparison, we report BOA evaluated on the entire 3DPW test set post adaptation (i.e. ``eval at the end"). Note that BOA is optimized for the streaming data setting. For that reason, they achieve decent results evaluating on streaming frames ($53$ Rec. Err on 3DPW-test), but their post-adaptation model ($79.18$ Rec. Err on 3DPW-test) overfits to the most recent frames. DAPA, on the other hand, optimizes a model on the entire target dataset and therefore is more generalizable after adaptation. Overall, we show that DAPA achieves better results than SPIN, EFT and the comparable version of BOA (``eval at the end").
\vspace{-\baselineskip}
\subsubsection{Finetuning experiments on SEEDLingS}
\begin{figure}
\centering
\begin{subfigure}{0.8\columnwidth}
     \includegraphics[width=\columnwidth]{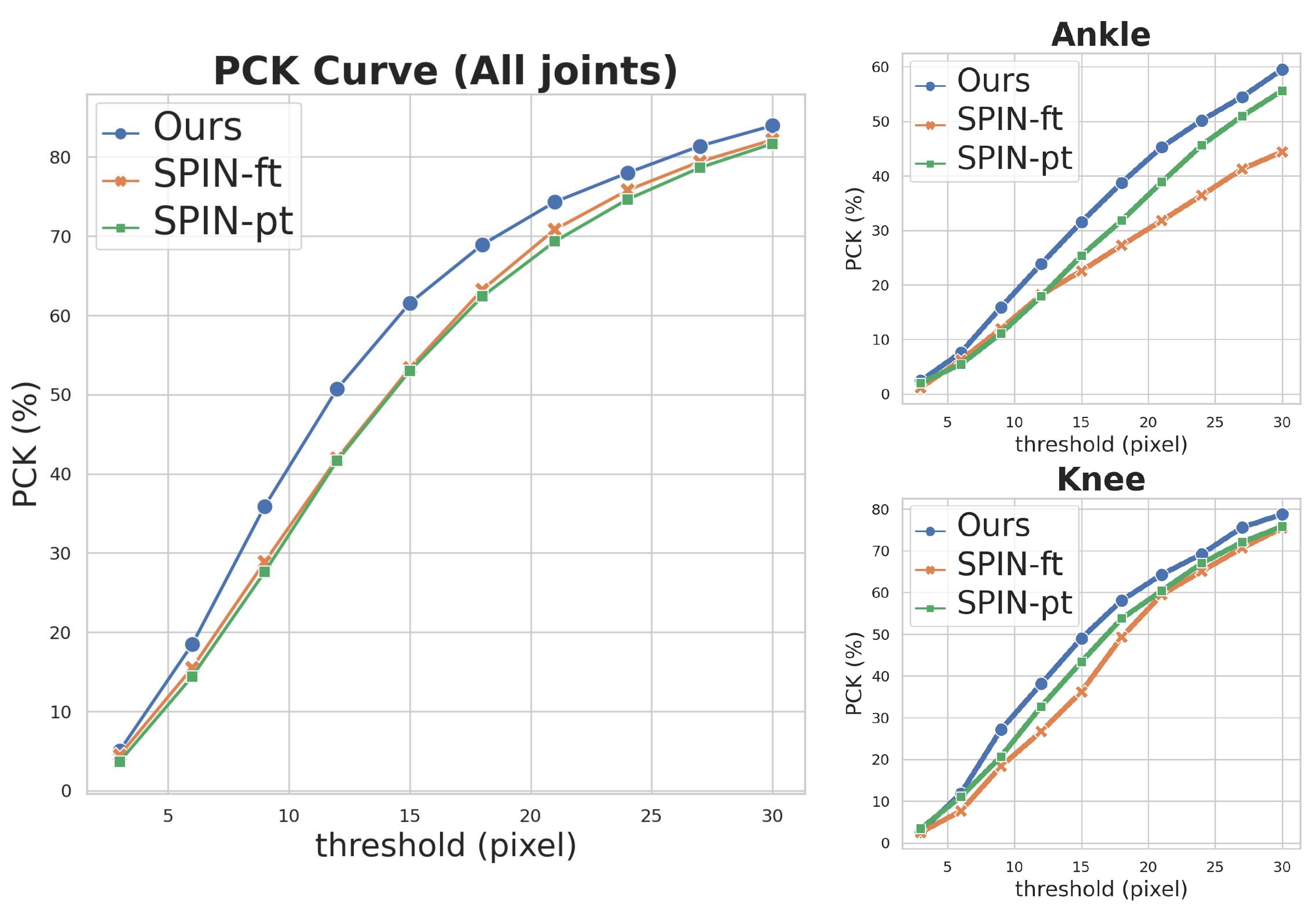}
     \caption{PCK Curves.}
     \label{fig:pck_curve}
 \end{subfigure}
 \hfill
 \begin{subfigure}{0.7\columnwidth}
     \includegraphics[width=\columnwidth, trim=0 0 0 20, clip]{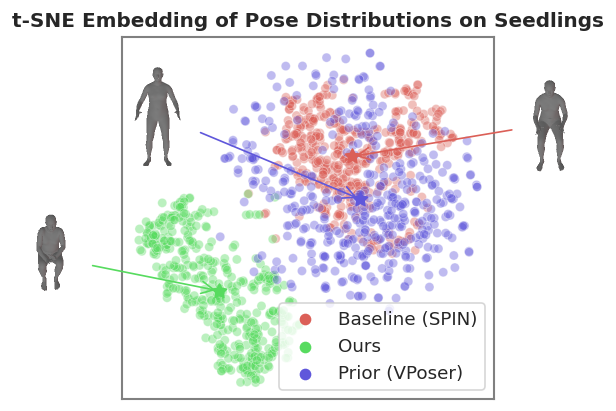}
     \caption{t-SNE visualization.}
     \label{fig:tsne}
\end{subfigure}
\caption{(a) PCK curves for SEEDLingS on all joints as well as ankles and knees. With different PCK thresholds, our method consistently outperforms baselines in localizing the keypoints. (b) t-SNE visualization of pose distributions.}
\end{figure}

Last but not least, we highlight DAPA's effectiveness in adapting to a very different pose distribution using a challenging in-the-wild dataset SEEDLingS \citep{bergelson2019day}. In Table \ref{table:seedlings_results}, model SPIN-pt is the model pretrained on the source. SPIN-ft and DAPA (Full model) are both SPIN-pt finetuned on SEEDLingS, but with SPIN's method and our method respectively. Since we only have 2D keypoint annotations on the test set, we adopt 2D keypoint metric Percentage of Correct Keyponts (PCK). We report PCK@0.2: percentage of correct keypoints using 0.2 $\times$ torso length as the distance threshold. We include the PCK curve (Figure \ref{fig:pck_curve}) using different threshold values and show that our method consistently outperforms the baseline. We additionally show PCK curves for the most challenging keypoints in SEEDLingS, ankles and knees. We can see that for the ankles, the PCK for SPIN-ft is even worse than SPIN-pt, since OpenPose often makes erroneous ankle predictions when the adults are kneeling. Therefore, SPIN-ft worsens the performance of SPIN-pt, by supervising the regression network with the intermediate SMPL bodies fitted to those noisy 2D keypoints. 
Table \ref{table:seedlings_results} contains quantitative results for ablated versions of DAPA: (1) training with randomly sampled poses from VPoser (``Rand. pose"); and (2) using the textures from SURREAL \cite{varol2017learning} in creating the synthetic images, rather than learned textures from the target dataset (``SURREAL textures"). Results suggest that the pose generation and texture learning in DAPA are both essential. 

\begin{table}[h]
\centering
\scriptsize
\resizebox{\columnwidth}{!}{\begin{tabular}{lc|ccccccc}
\toprule
\multicolumn{1}{c}{\multirow{2}{*}{Models}} & \multicolumn{7}{c}{PCK@0.2}   \\ 
 & all joints & eye & shoulder & elbow & wrist & hip & knee & ankle \\ \midrule
 SPIN-pt & 52.8 & 83.3 & 51.1 & 71.2 & 65.0 & 50.1 & 43.2 & 25.2 \\
 SPIN-ft & 53.1 & \textbf{91.1} & 62.2 & 73.9& \textbf{73.0} & 35.6 & 36.1 & 22.6 \\
 DAPA (Ours) \\
\;\;- Rand. pose & 53.2 & 81.0 & 56.1 & 73.0 & 64.0 & 53.6 & 41.7 & 26.2\\
\;\;- SURREAL & 57.7 & 83.3 & 74.8 & 65.7 & 65.7 & 62.8 & 44.5 & 28.0 \\
\;\;- Full model & \textbf{61.3} & 85.9 & \textbf{79.5} & \textbf{74.1} & 67.5 & \textbf{63.4} & \textbf{48.8} & \textbf{31.4} \\ \bottomrule
\end{tabular}}
\caption{Quantitative results on SEEDLingS. }
\label{table:seedlings_results}
\end{table}

\begin{figure*}[ht!]
\centering
\includegraphics[width=0.95\textwidth,page=1]{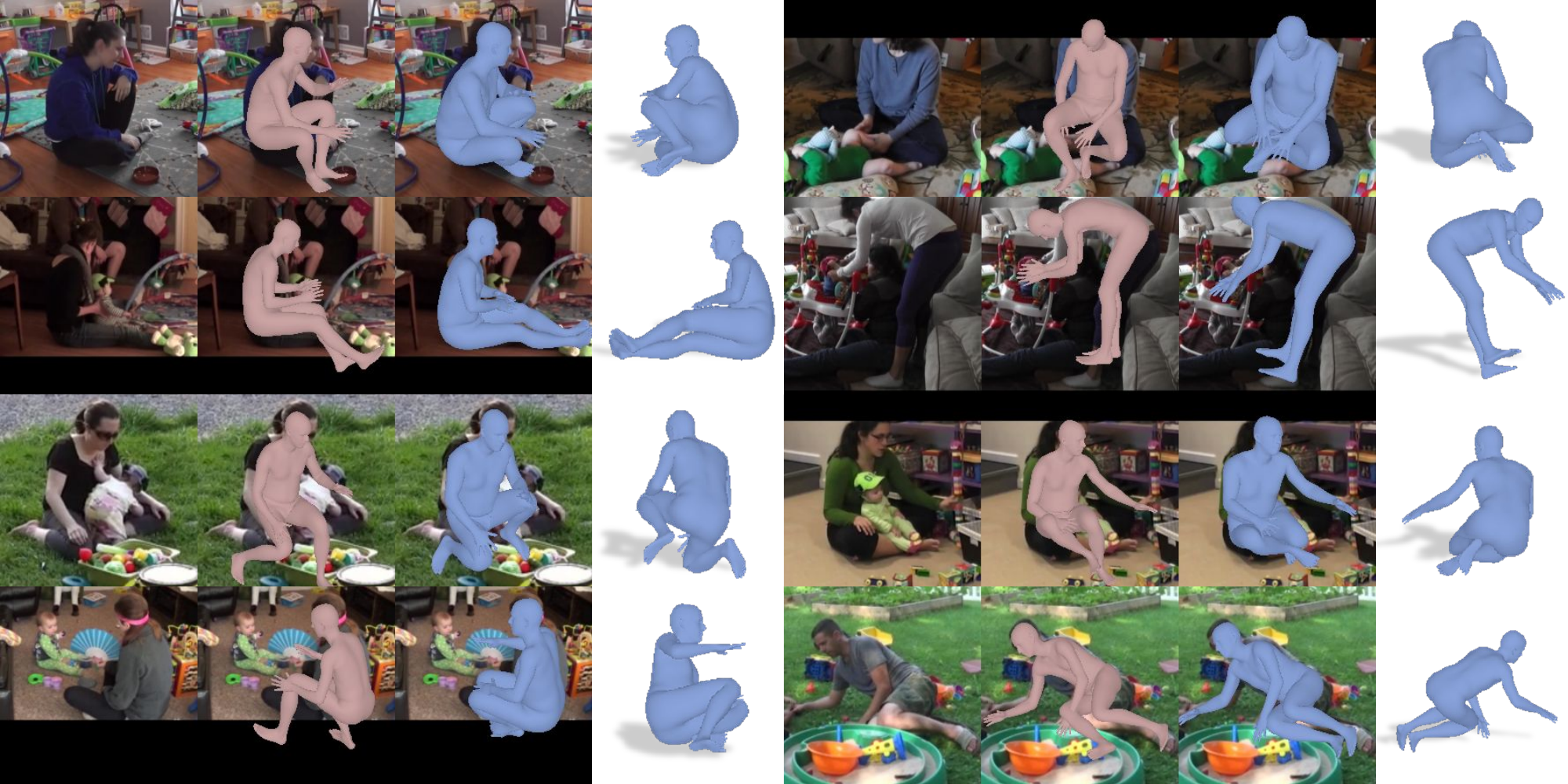}
\caption{Qualitative results on SEEDLingS. Each column, left to right: input, SPIN-ft result, DAPA (Ours), DAPA (Ours) (back view).}
\label{fig:qualitative}
\end{figure*}
\begin{figure*}[h!]
\centering
\includegraphics[width=0.95\textwidth]{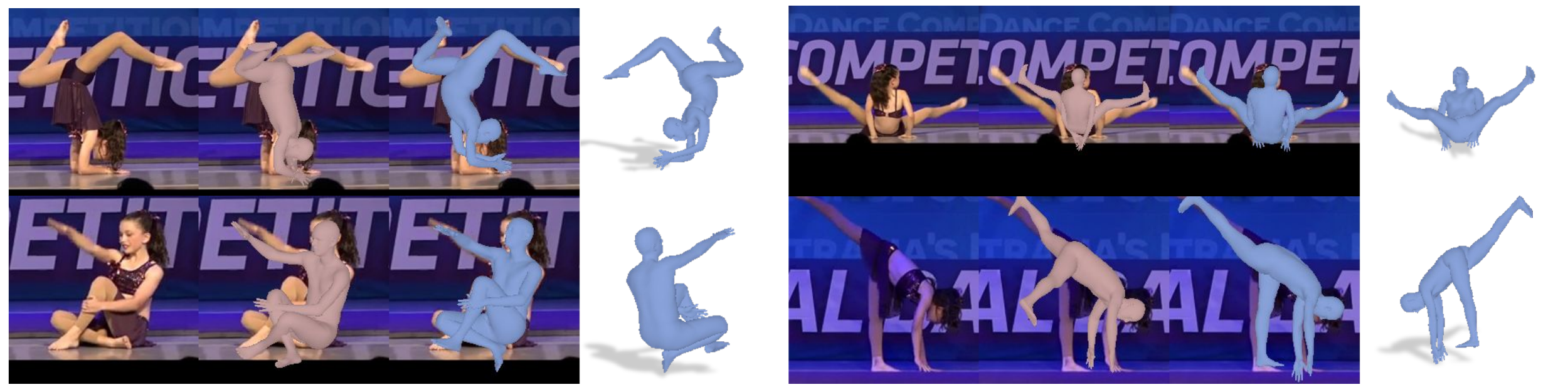}
\caption{Results on a YouTube gymnastics video with extreme poses. Each column, left to right: input, SPIN-ft result, DAPA result, DAPA result (back view).}
\label{fig:gymnastics}
\end{figure*}

Figure \ref{fig:tsne} visualizes the t-SNE \cite{van2008visualizing} embeddings of the predicted poses using SPIN-ft (in red) and our method (in green) in the VPoser latent space. We see that the predicted poses from SPIN-ft are close to the VPoser prior (i.e. normal distribution), whereas our method effectively shifts the predicted pose distribution to account for the specificity of the poses in SEEDLingS.

We compare SPIN-ft and DAPA qualitatively in Figure \ref{fig:qualitative}. DAPA generalizes better to the poses in SEEDLingS where the adults are often kneeling or sitting in various ways. SPIN-ft, on the other hand, tends to keep predicting poses that are closer to the canonical pose and fails to capture most of the kneeling poses. 

\vspace{-\baselineskip}
\subsubsection{Additional results on extreme poses}
To further demonstrate DAPA's utility, we finetune on a challenging YouTube gymnastics video \footnote{Youtube Video: \href{https://www.youtube.com/v/PSBOjqCtpEU}{https://www.youtube.com/v/PSBOjqCtpEU}} using OpenPose keypoints (Figure~\ref{fig:gymnastics}). We observe that DAPA is able to reconstruct extremely difficult acrobatic poses.
\vspace{-3px}





\section{Conclusion}
\label{sec:conclusion}
We proposed DAPA, a domain adaptive pose augmentation method that enhances model generalization on in-the-wild datasets where 3D ground truths are not available. We achieve this by automatically generating synthetic images adapted to target domains with posed SMPL bodies and learned textures. On 3D benchmarks 3DPW, AGORA and a real-world parent-child interaction dataset SEEDLingS, we showcase the effectiveness of our approach to improve generalization of human mesh recovery. 
\vspace{-\baselineskip}
\paragraph{Acknowledgements} This material is based upon work supported by the National Science Foundation under Grant No. 2026498, as well as a seed grant from the Institute for Human-Centered Artificial Intelligence (HAI) at Stanford University. K.-C. W. acknowledges the support of this work by the Wu Tsai Human Performance Alliance at Stanford University and the Joe and Clara Tsai Foundation. This work is also supported by the CONIX Research Center, one of six centers in JUMP, a Semiconductor Research Corporation (SRC) program sponsored by DARPA.
\section{Supplementary}
\subsection*{A. Additional qualitative results}
We include additional qualitative results from finetuning SEEDLingS in Figure \ref{fig:supp_seedlings}. We show that DAPA is able to reconstruct challenging poses such as cross-legged sitting poses (row 3 and 4 in the second column of Figure \ref{fig:supp_seedlings}).

Additional test finetuning results on YouTube gymnastics videos are in Figure \ref{fig:supp_gymnastics}. SPIN-ft and DAPA are both finetuned on the video frames using OpenPose detections as pseudo 2D ground truths, and then evaluated on the same set of video frames. Results suggest that DAPA is better at capturing small details (e.g. head in row 1 column 2, and left arm in row 2 column 1).

In Figure \ref{fig:supp_agora}, we include additional AGORA \cite{patel2021agora} qualitative results for the models in Table 1 in the main paper. SPIN-ft-EFT and SPIN-ft-AGORA are finetuned on EFT and AGORA respectively with 3D ground truths. SPIN-ft-AGORA-2D and DAPA (Ours) are both finetuned using only 2D ground truths. We can see that compared to SPIN-pt, SPIN-ft-EFT and SPIN-ft-AGORA-2D, our method DAPA achieves better reconstruction quality. In addition, DAPA achieves competitive results as compared to SPIN-ft-AGORA despite using less supervision, and sometimes even does better on recovering body details (e.g. the legs in row 1 and 2). 

\subsection*{B. Quality of the learned textures}
\begin{figure}[h!]
\centering
\includegraphics[width=\columnwidth]{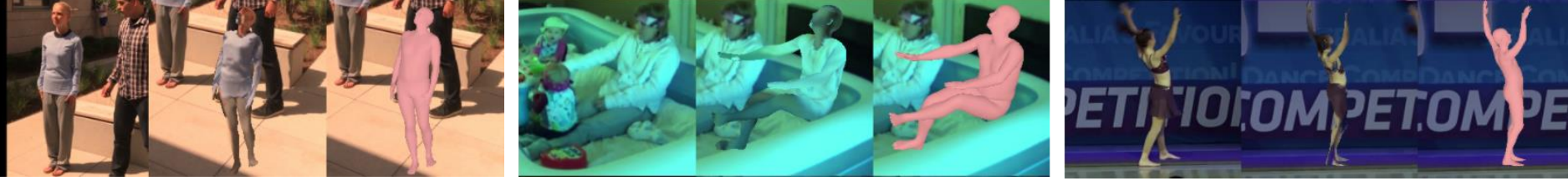}
\caption{Additional examples of synthetic examples. Left to right in each column: real image, synthetic image, ground truth mesh for the synthetic image.}
\label{fig:textures}
\end{figure}

We include additional synthetic examples in Figure~\ref{fig:textures}. While we can use a better texture model, we observe that improving the quality of the textures is not crucial for our task. This is because the texture model is for \textit{mining hard poses and letting the regressor predict those poses}.
In fact, we explored jointly optimizing the texture model during finetuning. While it improves textures, it increases training time without improving performance. Besides, we note that the texture quality in works such as \citet{rajasegaran2021tracking} and \citet{rajasegaran2022tracking} are also not that great either but they were sufficient in their respective tasks.

\subsection*{C. Quality of 2D keypoints}
\begin{figure}[h!]
\centering
\includegraphics[width=0.7\columnwidth]{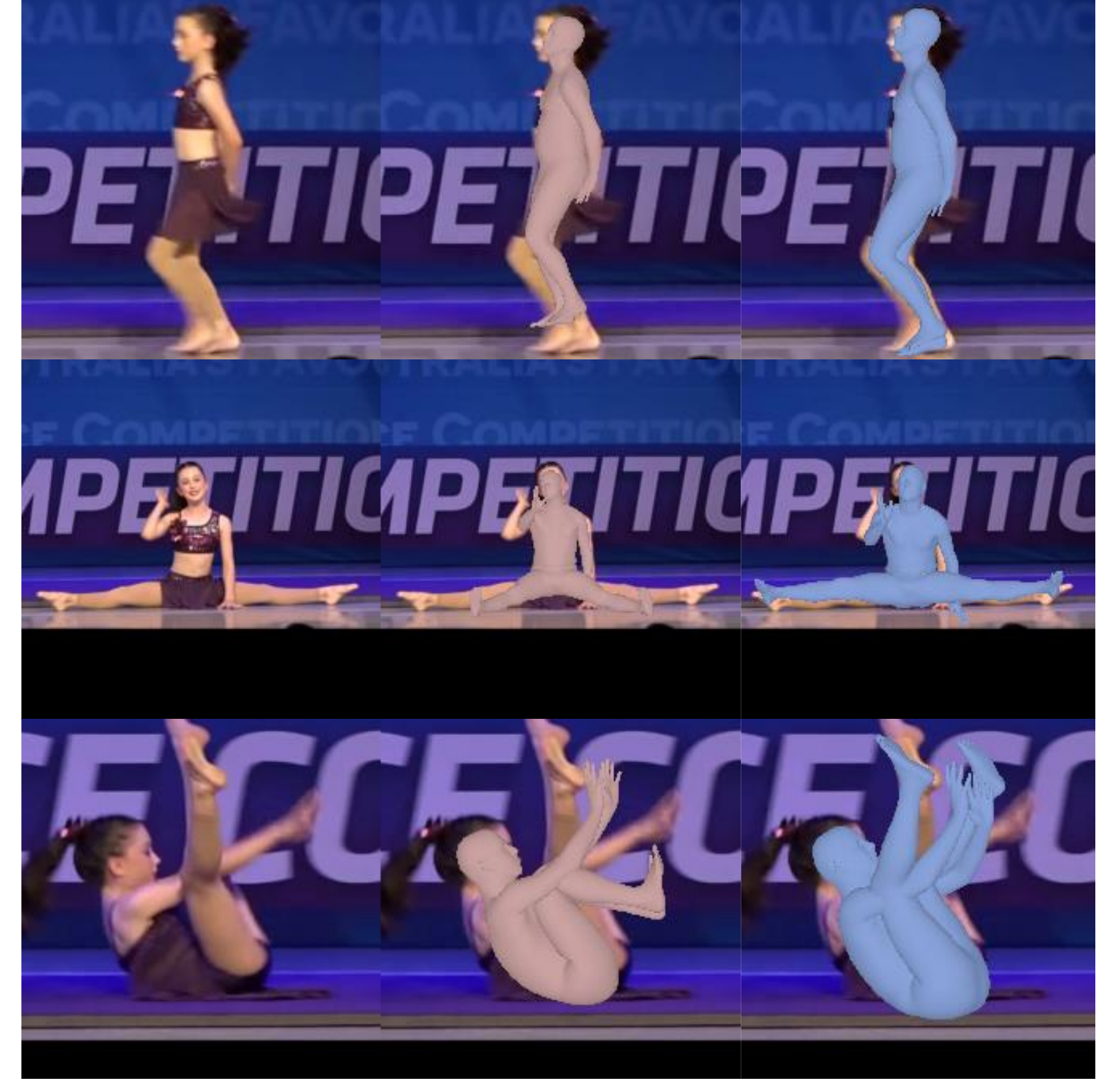}
\caption{From left to right: input image, DAPA trained with DarkPose \citep{zhang2020distribution}, DAPA trained with OpenPose.}
\label{fig:darkpose_comparison}
\end{figure}

In this work we use OpenPose as 2D keypoint detector because it is the go-to detector for works in Human Mesh Recovery. We provide qualitative comparison of DAPA using OpenPose versus a more recent keypoint detector \citep{zhang2020distribution} (Figure~\ref{fig:darkpose_comparison}). We observe that in general, DAPA trained with OpenPose keypoints have the same quality as using a more state-of-the-art keypoint detector, if not better.

\subsection*{D. Preprocessing details for SEEDLingS}
Here we describe the detailed preprocessing steps for constructing the SEEDLingS~\cite{bergelson2019day} datasets.

The SEEDLingS dataset contains hour-long recordings of 46 subjects. The videos contain the main view captured with a fixed camera in the room as well as head-mounted camera views. We crop the main view and resize it to $1920$ by $1080$ pixels. We run OpenPose with default settings to detect 2D keypoints in the frames. The training set is constructed by randomly sampling $50$ frames from each recording. 

The input to our model is 224 by 224 pixel bounding boxes centered on each human. We crop the humans using the OpenPose keypoints and scale the bounding boxes for each person such that the torso of the person is roughly one third of the box height. Additionally, we filter out persons with small bounding boxes before scaling as well as low confidence OpenPose detections, so that most infants are filtered out.

On the test set we have annotations for 14 keypoints per persons: (left and right) eyes, ears, shoulders, elbows, wrists, hips, knees, and ankles.

\subsection*{E. Additional experiment: finetune AGORA with OpenPose keypoints}
In the AGORA experiment in the main paper, we finetune using the ground truth 2D keypoints. Here we consider the setting where 2D ground truth annotations are not available during finetuning, in which case we use OpenPose \cite{cao2019openpose} detections to supervise the training. This is a representative setting of deploying pretrained human mesh recovery models on real-world data without any manual annotations. We can see that when finetuned with OpenPose keypoints, the performance of SPIN-ft drops. However, the performance of DAPA is better than the previous experiment where ground truth 2D keypoints were used. This is likely because that most heavily occluded humans are not detected by OpenPose and hence not included in finetuning. In the previous experiment, those extreme occluded humans may have introduced noise via the bad texture predictions. 

\subsection*{F. Limitations and Future Directions}
\begin{figure*}[h!]
\centering
\includegraphics[width=0.9\textwidth]{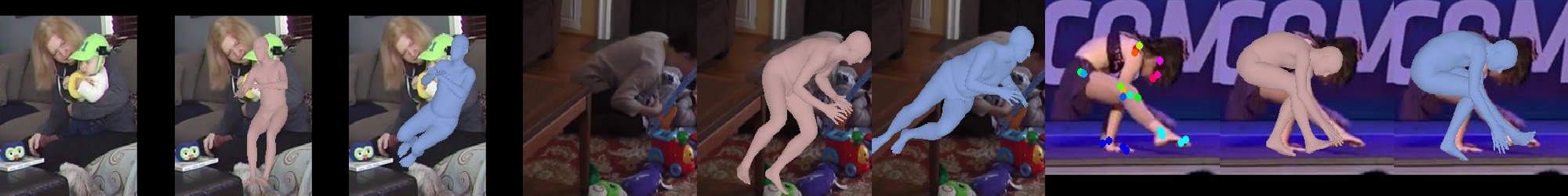}
\caption{Failure cases due to multiple persons (left), occlusion (middle) and erroneous 2D detections (right). The right column additionally shows OpenPose results, where arms are detected as legs (blue dots) by mistake. Each column, left to right: input image, SPIN-ft result, DAPA result.}

\label{fig:limitations}
\end{figure*}

\begin{figure*}[h]
\centering
\includegraphics[width=\textwidth]{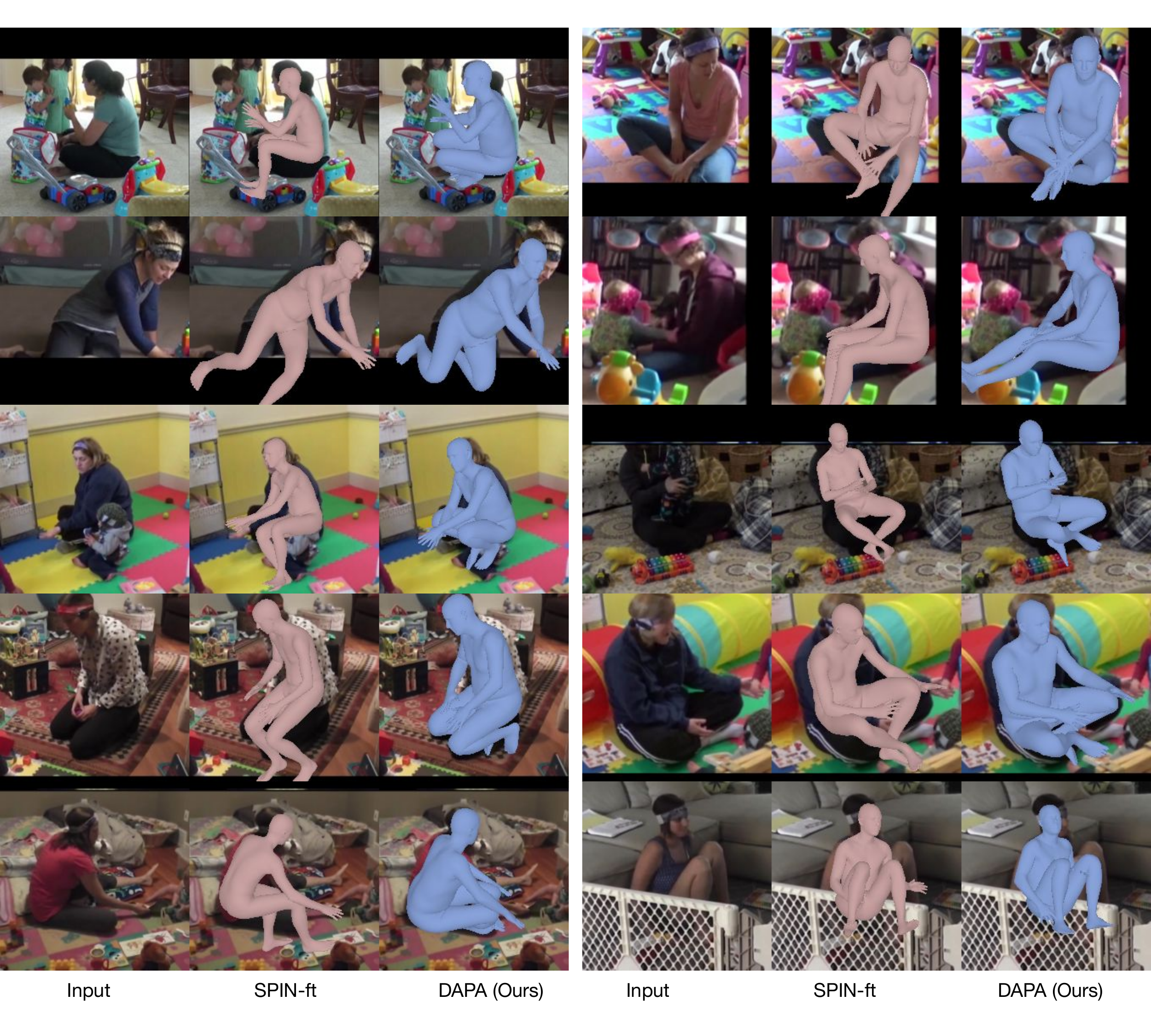}
\caption{Additional qualitative results on SEEDLingS. }
\label{fig:supp_seedlings}
\end{figure*}

\begin{figure*}
\centering
\includegraphics[width=\textwidth]{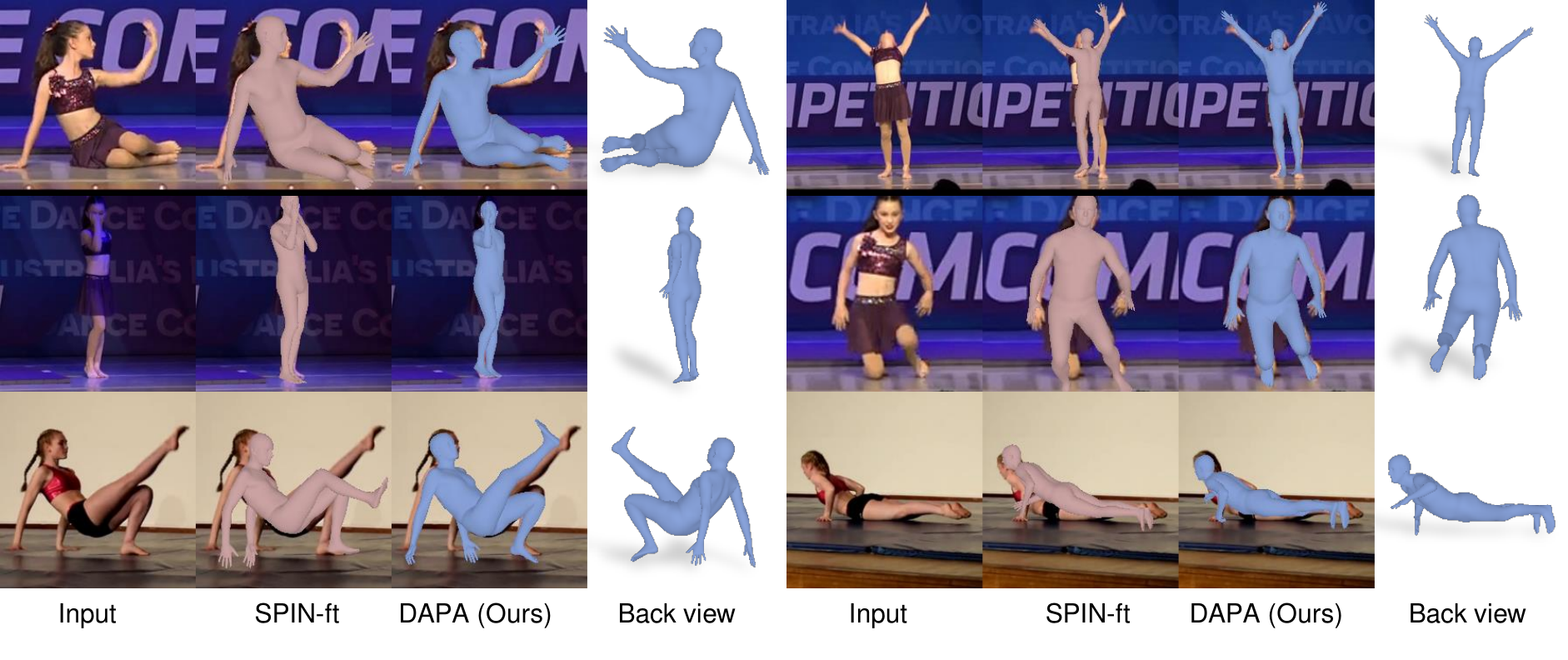}
\caption{Qualitative results on a gymnastics video. }
\label{fig:supp_gymnastics}
\end{figure*}

\begin{figure*}
\centering
\includegraphics[width=0.98\textwidth]{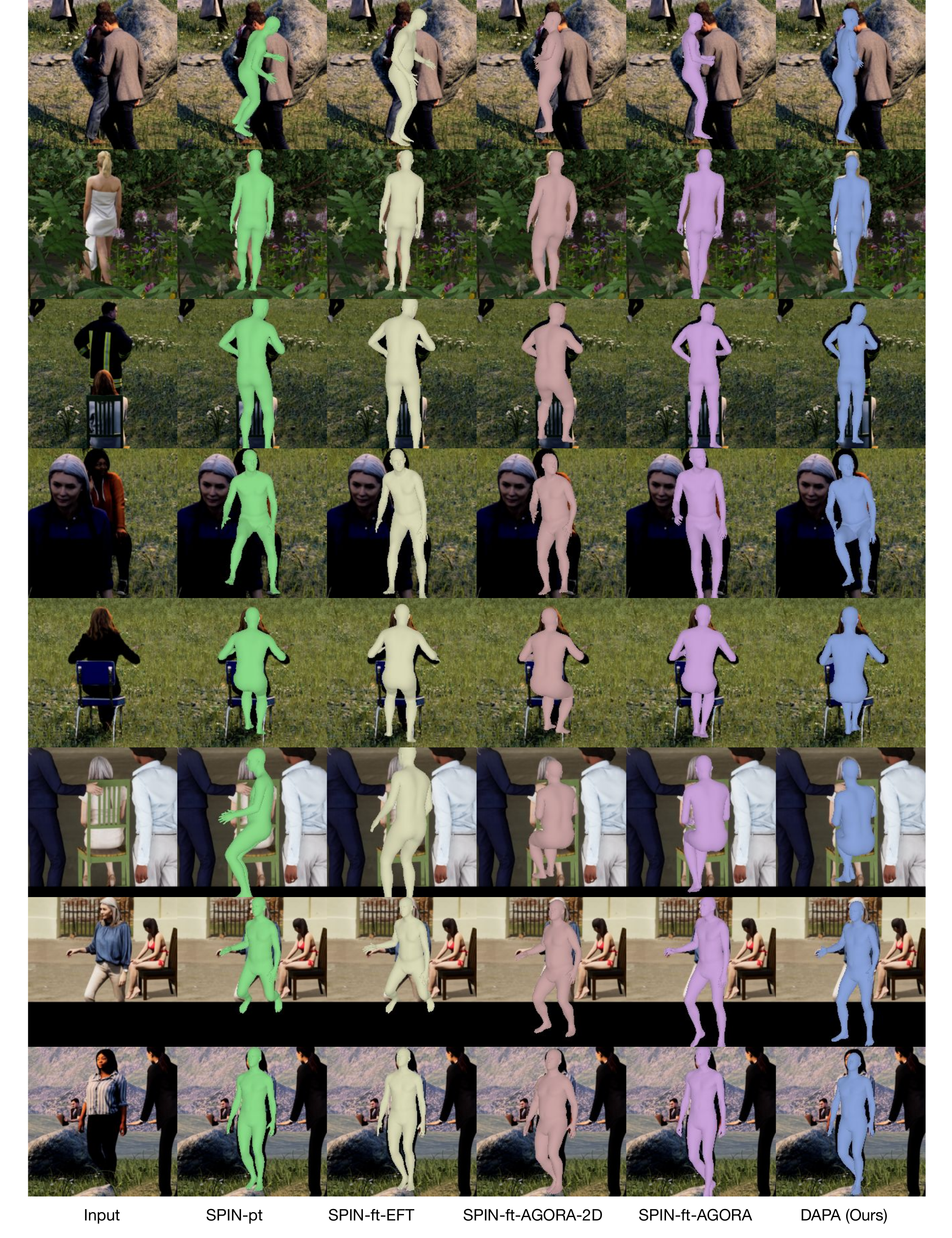}
\caption{Additional qualitative results on AGORA. }
\label{fig:supp_agora}
\end{figure*}

In Figure \ref{fig:limitations} we include example failure cases on SEEDLingS and the gymnastics video due to different human body shapes (e.g. infants) (left column), multiple persons in the frame (left column), pose ambiguity arising from heavy occlusion (middle column), and erroneous 2D detections (right column). Future work could consider extending DAPA to reconstruct multiple people or human bodies with diverse shapes such as infants.

\begin{table}[h]
\centering
\small
\resizebox{\columnwidth}{!}{
\begin{tabular}{@{} l *5c @{}}
\toprule
\multicolumn{1}{c}{Models} & MPJPE ($\downarrow$)  & MVE ($\downarrow$) & NMJE ($\downarrow$) & NMVE ($\downarrow$) & F1 ($\uparrow$) \\ 
\midrule 
 SPIN-ft &  210.3 & 202.4 & 269.6 & 259.5 & 0.78 \\
 DAPA (Ours) & \textbf{162.5} & \textbf{158.2} & \textbf{205.7} & \textbf{200.3} & \textbf{0.80}  \\ \bottomrule
\end{tabular}
}
\caption{\footnotesize Additional quantitative results on AGORA test set when no ground truth annotations are available during fintuning.}
\label{tab:my_label}
\end{table}

{\small
\bibliographystyle{abbrvnat}
\bibliography{egbib}
}

\end{document}